\documentclass[letterpaper, 10 pt, conference]{ieeeconf}  % Comment this line out if you need a4paper

\IEEEoverridecommandlockouts                              % This command is only needed if 
\usepackage{graphics} % for pdf, bitmapped graphics files
\graphicspath{ {./images/} }
\usepackage{subcaption}
\usepackage{ifthen}
\usepackage{amsmath} % assumes amsmath package installed
\usepackage{amssymb}  % assumes amsmath package installed
\usepackage{algorithm}
\usepackage[noend]{algpseudocode}
\usepackage{lipsum}
\usepackage{gensymb}
\usepackage{amsthm}
\usepackage{float}
\newtheoremstyle{mystyle}%                % Name
  {}%                                     % Space above
  {}%                                     % Space below
  {\itshape}%                                     % Body font
  {}%                                     % Indent amount
  {\bfseries}%                            % Theorem head font
  {.}%                                    % Punctuation after theorem head
  { }%                                    % Space after theorem head, ' ', or \newline
  {}%                                     % Theorem head spec (can be left empty, meaning `normal')
\theoremstyle{mystyle}
\newtheorem{definition}{Definition}
\newtheorem{constraint}{Constraint}
\newtheorem{assumption}{Assumption}
\usepackage{todonotes} % for missingfigure

\DeclareMathOperator*{\argmax}{arg\,max}

\title{\LARGE
Learning to Generate 6-DoF Grasp Poses with Reachability Awareness
}

\author{Xibai Lou$^{1}$, Yang Yang$^{2}$ and Changhyun Choi$^{1}$% <-this % stops a space
\thanks{*This work was in part supported by the MnDRIVE Initiative on Robotics, Sensors, and Advanced Manufacturing.}% <-this % stops a space
\thanks{$^{1}$X. Lou and C. Choi are with the Department of Electrical and Computer Engineering, Univ. of Minnesota, Minneapolis, USA
        {\tt\small \{lou00015, cchoi\}@umn.edu}}%
\thanks{$^{2}$Y. Yang is with the Department of Computer Science and Engineering, Univ. of Minnesota, Minneapolis, USA {\tt\small yang5276@umn.edu}}%
}

\begin{document}

\maketitle
\thispagestyle{empty}
\pagestyle{empty}

%%%%%%%%%%%%%%%%%%%%%%%%%%%%%%%%%%%%%%%%%%%%%%%%%%%%%%%%%%%%%%%%%%%%%%%%%%%%%%%%
\begin{abstract}
Motivated by the stringent requirements of unstructured real-world where a plethora of unknown objects reside in arbitrary locations of the surface, we propose a voxel-based deep 3D Convolutional Neural Network (3D CNN) that generates feasible 6-DoF grasp poses in unrestricted workspace with reachability awareness. Unlike the majority of works that predict if a proposed grasp pose within the restricted workspace will be successful solely based on grasp pose stability, our approach further learns a reachability predictor that evaluates if the grasp pose is reachable or not from robot's own experience. To avoid the laborious real training data collection, we exploit the power of simulation to train our networks on a large-scale synthetic dataset. This work is an early attempt that simultaneously evaluates grasping reachability from learned knowledge while proposing feasible grasp poses with 3D CNN. Experimental results in both simulation and real-world demonstrate that our approach outperforms several other methods and achieves 82.5\% grasping success rate on unknown objects.
\end{abstract}
\smallbreak

%%%%%%%%%%%%%%%%%%%%%%%%%%%%%%%%%%%%%%%%%%%%%%%%%%%%%%%%%%%%%%%%%%%%%%%%%%%%%%%%
\section{INTRODUCTION}
\begin{figure}[t]
\centering
    \includegraphics[scale=0.49]{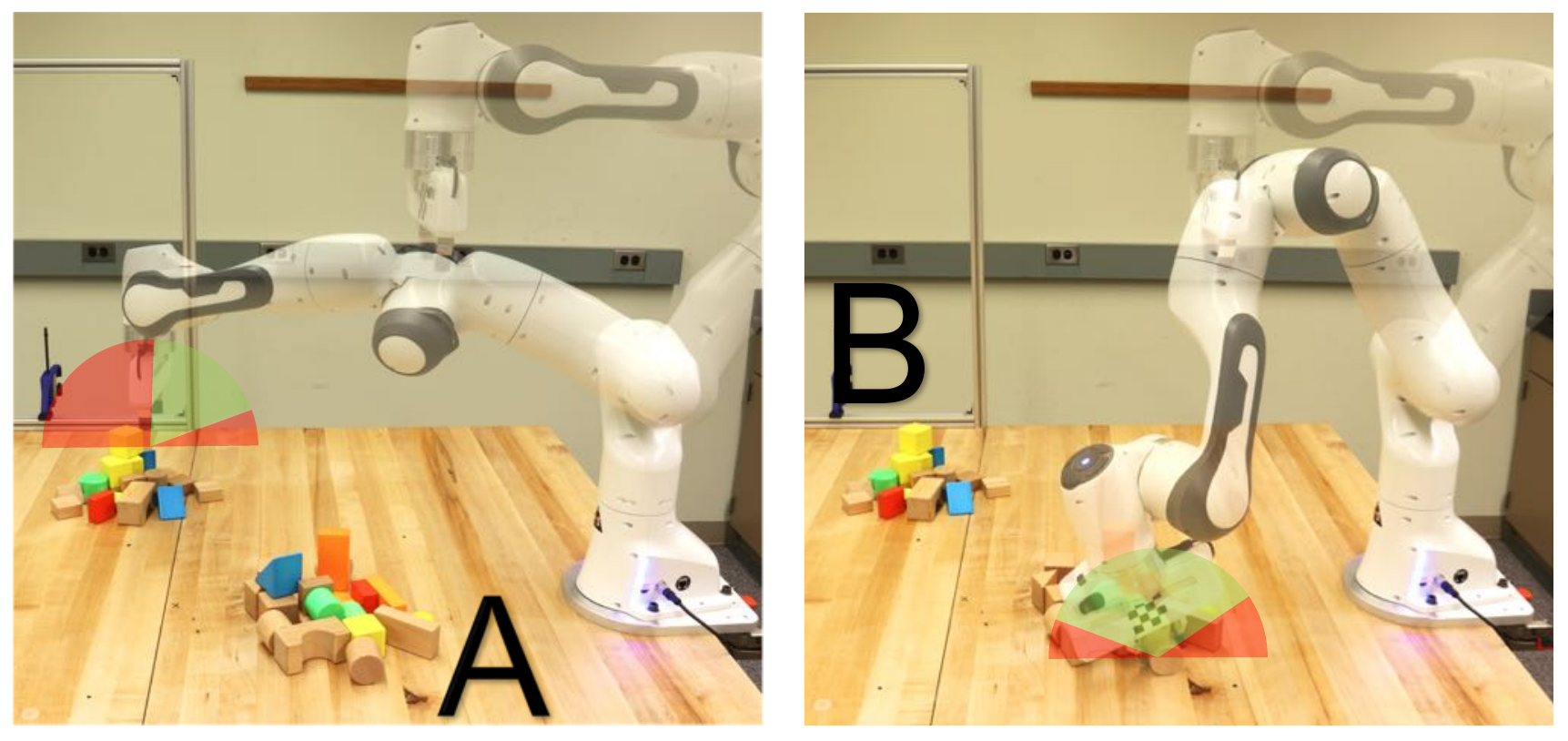}
\caption{\textbf{Example of searching for a feasible grasp pose.} Two clusters of objects are randomly arranged on the table. The green masks represent reachable approaching directions whereas the red mask is for unreachable ones. Note that in the left figure the robot has reached its limit, wherein the chance of finding a stable and reachable grasp in clutter A is much greater than that in clutter B.}
\label{fig:cover}
\vspace{-8pt}
\end{figure}

Real-world applications demand robotic manipulation algorithms that are efficient in arbitrary workspace where objects may not be reachable. Fig.~\ref{fig:cover} illustrates a scenario where such an algorithm needs to 1) decide which of the sampled grasp pose candidates are more reachable and 2) grasp as many objects as possible from the dense clutter with minimal efforts.

The predominant top-down grasping is often restricted in narrowly prepared workspace \cite{zeng2018learning}, whereas practical problems are often in extended and obstacle-rich environments that require flexible 6-DoF grasp poses to reach objects. Albeit extensive researches have been conducted on this topic, the grasping reachability problem remains relatively unexplored. Current 6-DoF approaches grasp within restricted workspace and only predict successful grasps by analyzing grasp poses and object shapes \cite{ten2017grasp}. When applied to unrestricted workspace, however, these approaches experience excessive planning failures that jeopardize grasping efficiency. We present a reachability aware 3D deep Convolutional Neural Network (3D CNN) that addresses these concerns by proposing feasible 6-DoF grasp poses that are both stable and reachable. 

Our approach consists of a 3D CNN and a Reachability Predictor (RP). 3D CNN learns spatial information \cite{maturana2015voxnet} which is effective in learning stable 6-DoF grasp poses and generalizing to novel objects \cite{choi2018learning}. Furthermore, the relatively small sim-to-real gap brought by depth is promising for direct real-world application. RP is effective in estimating the reachability of the sampled grasp poses without going through the computationally expensive motion planning algorithms. This attribute is jointly determined by the grasp pose and the kinematic constraints of a robot arm. Our approach discovers this intervened relationship and improves grasping efficiency by learning to approximate the grasping reachability from self-exploring experience. The immediate challenge here is how to train these models, which typically require hundreds of thousands of labeled data in order to generalize. Many learning-based grasping approaches suffered from insufficient training data since real robot data are notoriously expensive to collect. We exploit the power of a robot simulator and solve this problem with large-scale self-supervision.

Our work is inspired by human behavior; we naturally prefer to grasp closer objects with appropriate hand and arm poses, whereas for distant objects we either adjust our hand pose or abort grasping. Likewise, we propose an approach to mimic such a highly efficient grasping strategy. We conducted several experiments and ablation studies in simulation as well as real-world where our approach outperforms several comparable approaches in densely cluttered settings and generalizes to novel objects. To the best of our knowledge, our grasping strategy is the first attempt to generate reachability aware 6-DoF poses in dense clutter using a voxel-based 3D CNN. The main contributions of our work are bi-folded:

\begin{itemize}

\item 
\textbf{3D CNN-based grasp pose generation in 6-DoF} takes advantage of our grasp pose sampling algorithm that uniformly samples over the entire 3-dimensional space. In order to predict feasible 6-DoF grasp pose and generalize to novel objects, we exploit large-scale synthetic data collection via self-supervision. Furthermore, the domain-invariant nature of 3D CNN facilitates the direct application to real robot. 

\end{itemize}

\begin{itemize}

\item \textbf{Reachability Predictor} learns the robot capability from extensive self-exploration and eliminates the need for human-imposed constraints such as workspace restrictions and approaching direction filtering. It is able to predict the reachability of the sampled grasp pose candidates, thus increase the grasping and planning success rate of 6-DoF grasp pose in unrestricted workspace. Since it is decoupled from grasp learning, RP is adaptable to other robots.

\end{itemize}

\section{RELATED WORK}
\subsection{Object Grasping}

Though there are different taxonomies of robotic grasping \cite{sahbani2012overview}, \cite{bohg2013data}, the existing works of robotic grasping are commonly divided into two groups: traditional model-based and modern learning-based approaches.

Traditional model-based approaches often involve physical modeling of objects and thus require full knowledge of the objects such as shapes, weights, friction coefficients \cite{miller2004graspit}, \cite{ferrari1992planning}. More recent approaches utilize grasp quality metrics to select the force closure grasps from pre-planned sets by analyzing the contact wrench space \cite{weisz2012pose}. These approaches grasp efficiently with accurate measurements and models, however, the prerequisite efforts scale-up fast when implemented in real unstructured environment where novel objects are prevalent.

Recent learning-based approaches apply deep neural networks to robotic grasping \cite{lenz2015deep}, \cite{Kappler2015LeveragingBD}, \cite{pinto2016supersizing}, \cite{mahler2017dex}, \cite{ten2017grasp}, \cite{choi2018learning}, \cite{zeng2018learning}. The majority of these approaches employ convolutional neural networks (CNNs) that take monocular RGB images or 2.5D depth image as input and map the extracted features to a less complicated 3-DoF grasp pose that includes a grasping point on 2D image plane and a corresponding wrist orientation \cite{lenz2015deep}, \cite{pinto2016supersizing}, \cite{mahler2017dex}, \cite{zeng2018learning}, \cite{levine2018learning}. \cite{gualtieri2016high} and \cite{ten2017grasp} extend beyond the standard 3-DoF approaches. ten Pas et al. trained the networks with RGB-D images to evaluate a set of sampled grasp candidates. Since their sampling algorithm is based on geometric reasoning, they were able to find 6-DoF grasp poses. Due to the data hungry nature of learning-based approaches, generating large-scale training dataset is necessary. One approach is to label robot trails manually \cite{levine2018learning}, \cite{choi2018learning}. A less laborious way is to collect data in simulation \cite{pinto2016supersizing}, \cite{Kappler2015LeveragingBD}, \cite{mahler2017dex}. Although fast and scalable, the sim-to-real gap may render features learned in simulation inaccurate in real-world. Some re-train the network with real-world data \cite{zeng2018learning}, others solve the problem by either fine-tuning or domain adaptation \cite{fang2018learning}, \cite{chebotar2018closing}, \cite{bousmalis2018using}.

Originated from computer vision tasks such as object recognition \cite{maturana2015voxnet}, 3D data segmentation \cite{su2018splatnet}, and scene completion \cite{song2017semantic}, voxel-based 3D CNN have also been applied to robotic grasping \cite{choi2018learning}. Choi et al. showed 3D CNN trained with real-robot data is effective in classifying discrete grasp poses of a soft hand for a single object. Our works differ in two ways. First, our problem is much more difficult in that we search in a 6-DoF space, which is necessary for non-compliant grippers. Second, generalization of 6-DoF grasping requires exponentially more data that are only possible to collect in simulation, where we develop a data collection framework and show our approach can be directly applied to real robot.

\subsection{Grasp Planning}

Robotic manipulation such as grasping depends on solving an inverse kinematic problem to find a path for the manipulation tool. Modern approaches employ fast motion planning algorithms such as RRT \cite{LaValle00rapidly-exploringrandom}, and its variants \cite{kuffner2000rrt} . Traditionally the grasping range of a given robot arm is determined by its maximum manipulability of the closed kinematic chain \cite{park1998manipulability}. This problem is not trivial and an analytical solution is often not easy to find. Furthermore, due to the random exploring nature of the algorithms, a solution for valid grasp poses is not guaranteed. In practice, the majority of approaches in robotic grasping restricts testing objects in a reachable but restricted workspace \cite{ten2017grasp}, \cite{choi2018learning}, \cite{zeng2018learning}, forcing it to operate in a reduced capability. Our work is one of the early explorations of learning grasping reachability, which allows the robot to work at full capacity.

\section{PROBLEM FORMULATION}

\begin{figure*}[t]
\centering
    \includegraphics[width=\textwidth]{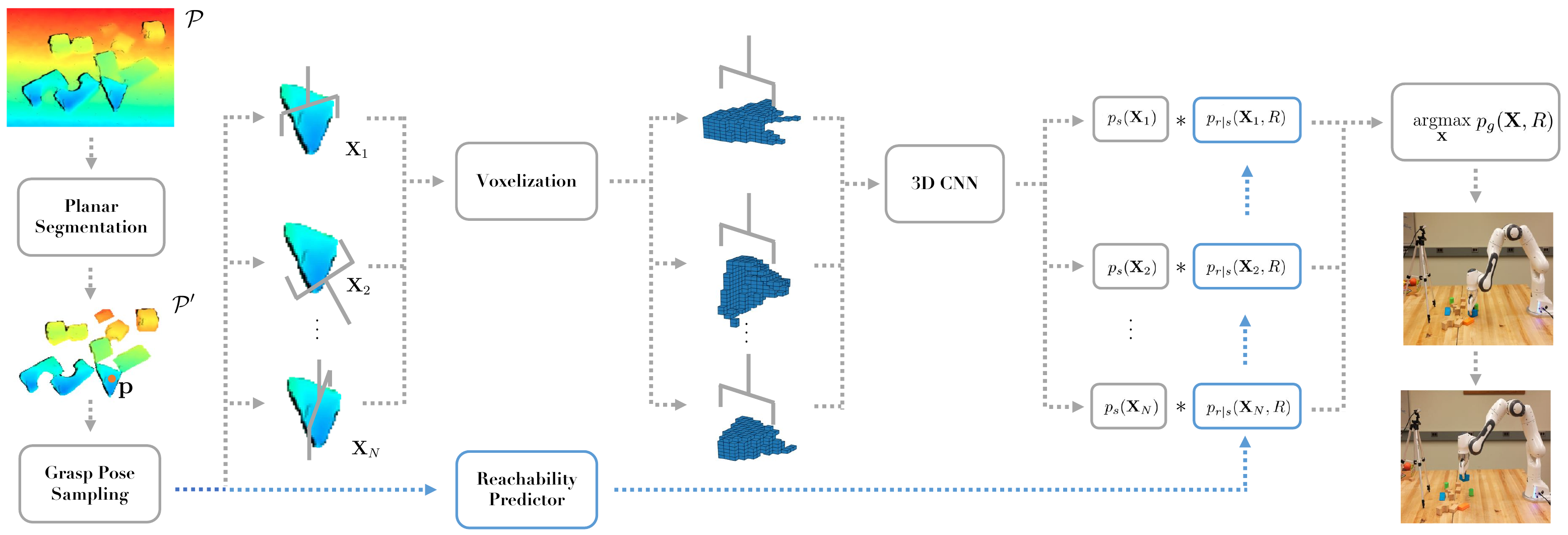}
\caption{\textbf{Grasping Pipeline.} Object point cloud $\mathcal{P'}$ is obtained from planar segmentation of point cloud $\mathcal{P}$. For each of the sampled grasp poses $\mathbf{X} \in \mathcal{X}$, the object point cloud $\mathcal{P'}$ is voxelized to voxel grid $\mathcal{V}$ and transformed by the corresponding grasp pose candidate $\mathbf{X}$. The input voxel grid is then passed to 3D CNN while the grasp candidate $\mathbf{X}$ is fed to RP for evaluation. The most probable grasp pose is chosen and executed by the robot manipulator.}
\label{fig:flowchart}
\vspace{-8pt}
\end{figure*}

We aim to find feasible 6-DoF grasp poses that are simultaneously stable and reachable. The problem is carried out in two stages. At the first stage, the robot finds a set of stable grasp candidates, some of which may not have valid motion plans due to invalid inverse-kinematic solution. At the second stage, the robot evaluates each grasp pose and excludes the unreachable ones. The problem can be formalized as follows:

\begin{definition}
\textit{A grasp pose $\mathbf{X}\in SE(3)$ is \textbf{stable} if it is able to form a force closure grasp. This attribute is independent of kinematic constraints.}
\end{definition}

\begin{definition}
\textit{A grasp pose $\mathbf{X}\in SE(3)$ is \textbf{reachable} if the given robot arm is able to achieve the pose without violating the physical limits of itself and the environment.}
\end{definition}

\begin{definition}
\textit{Given a point cloud $\mathcal{P} \subset \mathbb{R}^3$, the goal is to find a \textbf{feasible} grasp pose $\mathbf{X_f}\in SE(3)$ that is both \textbf{stable} and \textbf{reachable}}.
\end{definition}

The grasp pose $\mathbf{X}$ is defined with respect to the robot coordinate frame. The point cloud $\mathcal{P}$ is obtained via a depth sensor with known extrinsic parameters, by which the cloud $\mathcal{P}$ is transformed from the sensor coordinate frame to the robot coordinate frame. An important assumption is:

\begin{assumption}
\textit{The set of 6-DoF grasp candidates $\mathcal{X}$ is randomly generated over the points $\mathbf{p}$ in the point cloud $\mathcal{P}$, i.e., $\mathbf{p} \in \mathcal{P}$.}
\end{assumption}

Given a point cloud $\mathcal{P}$ and a grasp candidate $\mathbf{X} \in \mathcal{X}$, where $\mathcal{X}$ denotes a set of $N$ uniformly generated 6-DoF grasp candidates, let $\mathcal{S}_s(\mathbf{X})\in$ \{0,1\} denote a binary-valued stability metric where $\mathcal{S}_s = 1$ indicates that the grasp is stable according to Definition 1. Our goal is to estimate the grasping stability $p_s(\mathbf{X}) = Pr(\mathcal{S}_s = 1|\mathbf{X})$ by self-supervised learning. To train the network more efficiently, we constrain our grasp sampling algorithm as follows:

\begin{constraint}
\textit{The grasp pose $\mathbf{X}\in SE(3)$ is constrained in that the sampled grasp candidates are limited so as to approach the target object from the top hemisphere.}
\end{constraint}

For an arbitrary grasp candidate $\mathbf{X}$, the wrist orientation has no influence over its reachability since the joint limit is $(-\pi, \pi)$. Therefore, a valid robot grasp $\mathbf{X}$ is determined by the combination of grasp location $(x, y, z) \in \mathbb{R}^3$ and approaching direction $(a_x, a_y, a_z) \in \mathbb{R}^3$. We construct the reachability determinant $\mathbf{a} = (x, y, z, a_x, a_y, a_z)$. To allow the robot fully explore the workspace, we assume that:

\begin{assumption}
\textit{The robot workspace is unrestricted and can be anywhere within the camera observation space.}
\end{assumption}

Let $R$ denote the robot and $\mathcal{S}_r(\mathbf{X}, R) \in$ \{0,1\} a binary-valued reachability metric, where $\mathcal{S}_r = 1$ indicates that the grasp is reachable. We wish to learn to predict the grasping reachability $p_r(\mathbf{X}, R) = Pr(\mathcal{S}_r = 1|\mathbf{X}, R)$ from self-supervised exploration. It is important to notice that the stability metric of a pose $\mathbf{X}$ is independent of its reachability metric, i.e., $\mathcal{S}_s = 1$ does not indicate $\mathcal{S}_r = 1$ and vice versa.

\section{PROPOSED APPROACH}

\subsection{Generating Feasible Grasp Poses}

The objective of our approach is to predict the most feasible grasp pose from a set of randomly sampled candidates in multiple settings. According to Definition 3, the selected grasp pose should be both stable and reachable. Training one generic model on this task leads to unsatisfactory performance, as the network may falsely ascribe the reason of a failed grasp to unstable grasp poses, while the true cause is poor reachability, and vice versa. As mentioned in the previous section, these two prerequisites of a feasible grasp pose is entirely independent of each other. This allows us to solve the credit assignment problem by decoupling the task into two independent sub-problems. The grasping success probability $p_g(\mathbf{X},R)$ can be decomposed as follow:
\begin{equation*}\label{eq:pareto mle2}
\begin{aligned}
p_g &= Pr(S_g = 1|\mathbf{X}, R)\\
&= Pr(S_r = 1, S_s = 1|\mathbf{X}, R)\\
&= Pr(S_r = 1|S_s = 1, \mathbf{X}, R) \times Pr(S_s = 1|\mathbf{X})\\
&= p_{r|s}(\mathbf{X}, R) \times p_s(\mathbf{X})
\end{aligned}
\end{equation*}

We trained a 3D CNN grasp pose predictor and a reachability predictor to estimate the resulting grasping stability $p_s$ and grasping reachability $p_{r|s}$ respectively. 

The grasping pipeline is described in Fig.~\ref{fig:flowchart}. The system first obtains the object point cloud $\mathcal{P'}$ from planar segmentation of point cloud $\mathcal{P}$ and samples $N$ grasp candidates over $\mathcal{P'}$ via the sampling algorithm described in Section~\ref{section:sampling}. For each sampled grasp candidate, objects point cloud is voxelized to a 3D voxel  grid $\mathcal{V} \in \mathbb{Z}^{32 \times 32 \times 32}$, where each voxel in the  grid is either 0 (not occupied) or 1 (occupied). The total physical edge length of the voxel grid is $0.1m$, equivalent to crop the object point cloud by a $0.001m^3$ cubic box that centered at grasping point $\mathbf{p}$. We choose cropping rather than segmenting an object from the point cloud as it preserves surrounding geometry that helps avoid collisions in dense clutter. The object voxel grid $\mathcal{V}$ may partially contain voxels of adjacent objects, which contributes to lower grasping stability predictions. Therefore, less surrounded objects, i.e., objects on the peripherals, are prioritized, resulting in an onion-peeling grasping pattern. 

Our 3D CNN architecture is similar to those in \cite{maturana2015voxnet} and \cite{choi2018learning}. We embed the grasp pose within the voxel grid by transforming it with grasp candidate $\mathbf{X}$. The 3D CNN will predict the grasping stability $p_s(\mathbf{X})$ based on the transformed voxel grid. The reachability predictor then extracts the reachability determinant $\mathbf{a} \in \mathbb{R}^6$ from each randomly generated pose $\mathbf{X}$ and estimates the grasping reachability $p_{r|s}(\mathbf{X},R)$. The grasping success probability $p_g$ is calculated by multiplying the two terms, Algorithm 1 explains this prediction procedure in detail.

\begin{algorithm}[t]
\caption{Reachability Aware 3D CNN Grasping}\label{euclid}
\hspace*{\algorithmicindent} \textbf{Input: }point cloud $\mathcal{P}$, 3D CNN Grasp model $\mathcal{N}_s$, reachability predictor $\mathcal{N}_r$\\
\hspace*{\algorithmicindent} \textbf{Output: }feasible grasp pose $\mathbf{X_f}\in{SE(3)}$ 
\begin{algorithmic}[1]
\State $\mathcal{P'} \gets \texttt{PlanarSegmentation($\mathcal{P}$)}$
\State $\mathcal{X} \gets \texttt{GraspPoseSampling($\mathcal{P'}$)}$
\For{${\mathbf{X}}\in\mathcal{X}$}
    \State $\mathbf{a} \gets \texttt{Extract}(\mathbf{X})$
    \State $\mathcal{V} \gets \texttt{Voxelization}(\mathcal{P'},\mathbf{X})$
    % \State $\mathcal{V'} \gets \texttt{VoxelTransformation}(\mathcal{V}, \mathbf{X})$
    \State $p_{r|s} \gets \texttt{$\mathcal{N}_r$.Feedforward}(\mathbf{a})$
    \State $p_s \gets \texttt{$\mathcal{N}_s$.Evaluate($\mathcal{V}$)}$
    \State $p_g(\mathbf{X}) \gets p_{r|s} \times p_s$
    % \State $\mathbf{p}_g \gets p_a \cup \mathbf{p}_g$
\EndFor
\State $\mathbf{X_f} \gets \argmax_{\mathbf{X}} p_g(\mathbf{X})$
\State $\texttt{Grasp}(\mathbf{X_f})$
\end{algorithmic}
\end{algorithm}

\begin{figure}[t]
  \centering
    \includegraphics[scale=0.25]{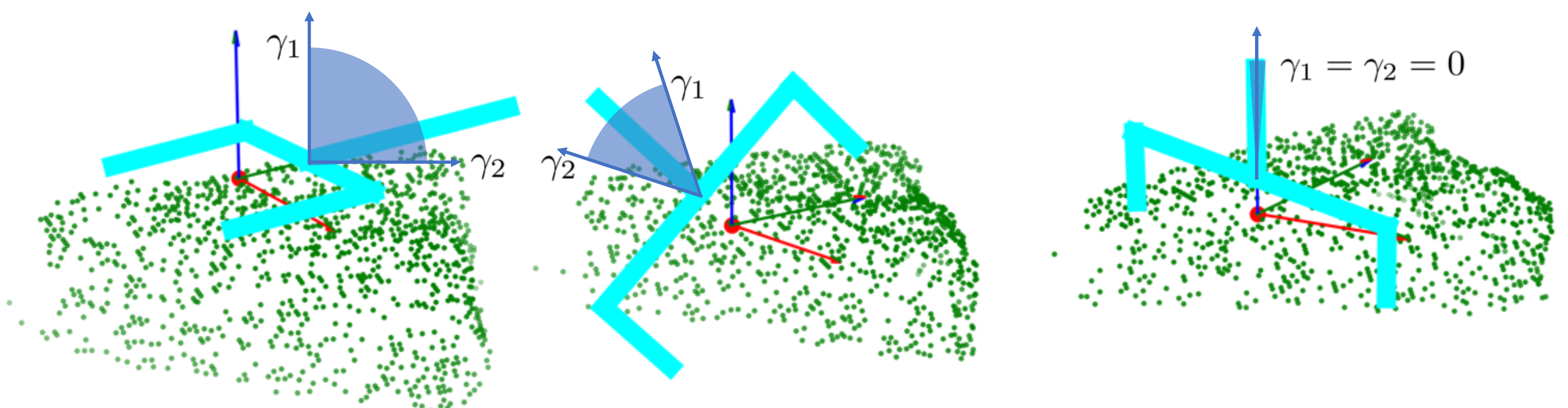}%
    \vspace{-3pt}
      \caption{\textbf{Examples of randomly sampled grasp candidates}. The grasp candidates are sampled over a triangle shape. For each candidate, the wrist orientation is sampled in $[0, 2\pi]$ while approaching direction in $[\gamma_1,\gamma_2]$. Different thresholds of the approaching direction are visualized here. Left figure is searching in full 6-DoF by setting $\gamma_1$ = 0\degree and $\gamma_2$ = 90\degree, the middle one shows a search constrained by $\gamma_1$ = 30\degree and $\gamma_2 = 60\degree$ and on the right is a top grasp pose by setting both $\gamma_1$ and $\gamma_2$ to 0.}
      \label{fig:graspsampling}
      \vspace{-8pt}
\end{figure}

\subsection{Grasp Pose Sampling Algorithm}
\label{section:sampling}
Unlike the geometric reasoning-based sampling algorithm proposed in \cite{ten2017grasp}, to ensure that the robot comprehensively explores the 6-DoF action space, we present a flexible algorithm that uniformly samples grasp candidates over the target objects within two threshold values. Given a desired number of samples $N$ and approaching vector thresholds $\gamma_1, \gamma_2 \in [0\degree,90\degree]$, where $\gamma_1 < \gamma_2$, the algorithm uniformly selects $N$ grasp points from the input point cloud $\mathcal{P}$. For each grasp point, a random pose is associated. Fig.~\ref{fig:graspsampling} shows three examples of the sampled grasp pose candidates. The sampling algorithm gives little constraints to the generated grasp candidates, so the network is able to learn the reasons of grasp failures from the unsuccessful data, such as grasping a corner or colliding with the object.

\subsection{Network Architecture}
We borrow the network structure from \cite{choi2018learning}; the first layer has 32 filters of size $5 \times 5 \times 5$, the second layer has 32 filters of $3 \times 3 \times 3$. The features are condensed by a Max Pooling layer of $2 \times 2 \times 2$, followed by two dense layers of 128 and 1. A given grasp pose can either be 1 (stable grasp) or 0 (unstable grasp), so we use binary cross-entropy as the loss function. We use the sigmoid activation function in the final layer to predict the grasping stability for the voxel grid $\mathcal{V}$.

Reachability predictor consists of an input layer, one hidden layer of size 16, one hidden layer of size 8 and a final output layer. The input to the neural network is a 6-dimensional reachability determinant $\mathbf{a}=(x, y, z, a_x, a_y, a_z)$, and the output is a binary classification result, where 1 denotes the grasp pose is reachable.

To integrate these two networks, we embed the grasping reachability into the grasping success probability by multiplying the two results. The final output $p_g$ indicates the grasping success probability of $\mathbf{X}$.

\subsection{Data Collection and Training}
\begin{figure}[t]
  \centering
    \includegraphics[scale=0.55]{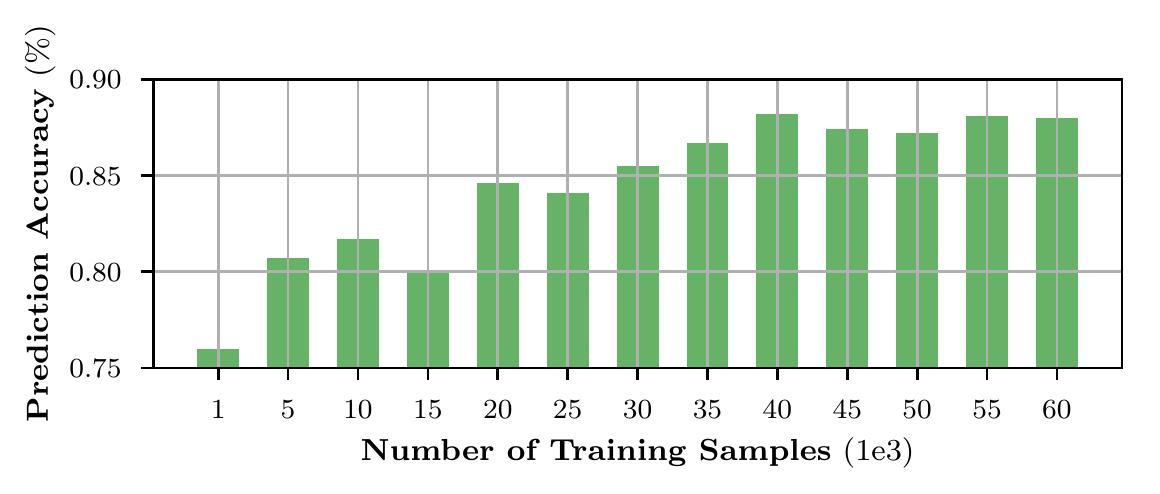}%
    \vspace{-8pt}
      \caption{\textbf{Prediction accuracy w.r.t. the size of dataset.} 3D CNN demands large-scale labeled data, tailored to the task, and such requirement is only attainable in simulation.}
      \vspace{-8pt}
      \label{fig:training}
\end{figure}
\begin{figure}[t]
    \centering
    \includegraphics[scale=0.38]{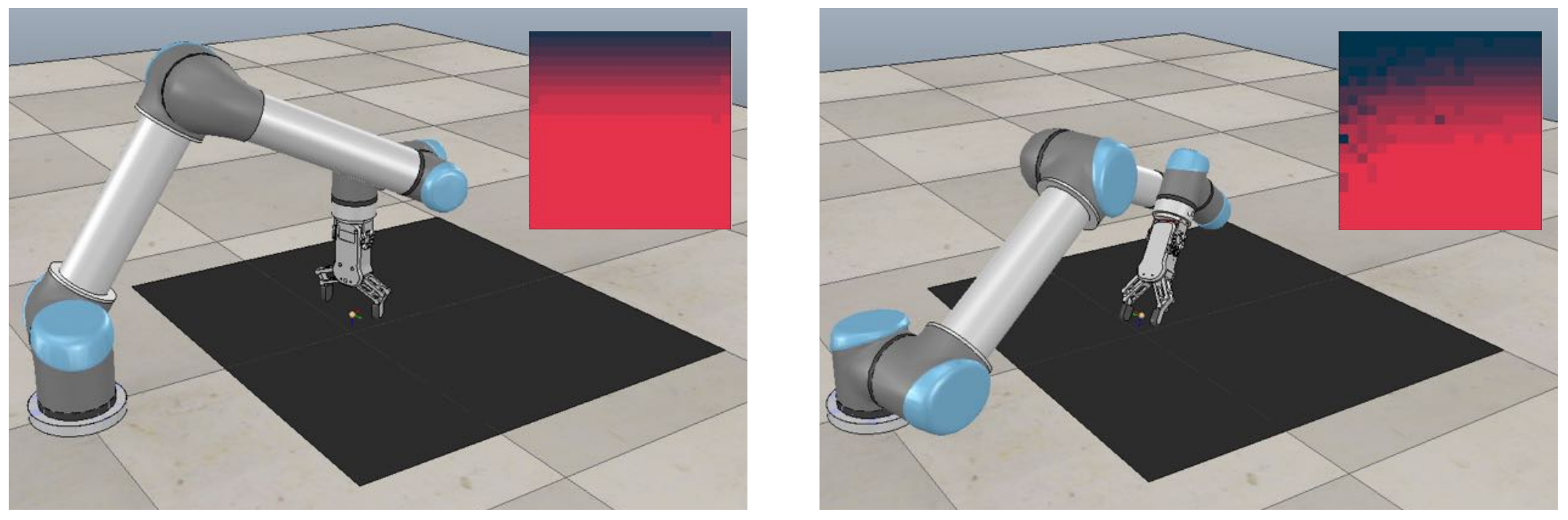}%
    \vspace{-5pt}
   \caption{\textbf{Visualization of Reachability Predictor}. We visualize the predictions for two poses over an arbitrary workspace, colored in black. Dark blue indicates poor reachability.}
%   \vspace{-3pt}
   \label{fig:rpviz}
\end{figure}

Training 3D CNN with multiple objects is ineffective because their shape varies for every grasp pose, preventing 3D CNN from generalizing object geometries. We used 8 primitive shapes from \cite{zeng2018learning} as our training objects. The self-supervised robot interacts with a single object and collects 60,000 labeled voxel grids to train the 3D CNN and 10,000 data to train the RP. A testing dataset of 1000 data is reserved to evaluate the prediction accuracy of 3D CNN. Fig.~\ref{fig:training} compiles this result with respect to the training data size from 1,000 to 60,000. It is clear that the prediction accuracy benefits from large-scale training data. We also show two prediction results of the RP in Fig.~\ref{fig:rpviz}, as expected, an approaching direction toward the robot itself results in narrower reachable space. Our approach learns to select candidates with more vertical approaching direction as they are generally more stable and reachable in our training and testing environments.

\section{EXPERIMENTS}
Our experiments aim to answer the following four questions: i) How much does our approach outperform other approaches? ii) Can our approach generalize to cluttered objects and novel objects? iii) Is the reachability predictor effectively targeting reachable objects? iv) Do our networks trained in simulation work well in real robot? 

To answer these questions, we conducted experiments in both simulated and real settings. We compare our approach with 3 other approaches: 1) \textbf{\texttt{RAND}}, a baseline that randomly generates a grasp pose without any learning, 2) \textbf{\texttt{GPD}}, Grasp Pose Detection proposes 6-DoF grasp candidates based-on geometric reasoning and evaluates them with RGB-D images trained CNN \cite{ten2017grasp} and 3) \textbf{\texttt{VPG}}, Visual Pushing for Grasping \cite{zeng2018learning} learns the synergy between pushing and grasping with a reinforcement learning to clean cluttered objects.

\begin{figure}[t]
   \centering
   \begin{subfigure}{0.15\textwidth}
       \includegraphics[scale = 0.32]{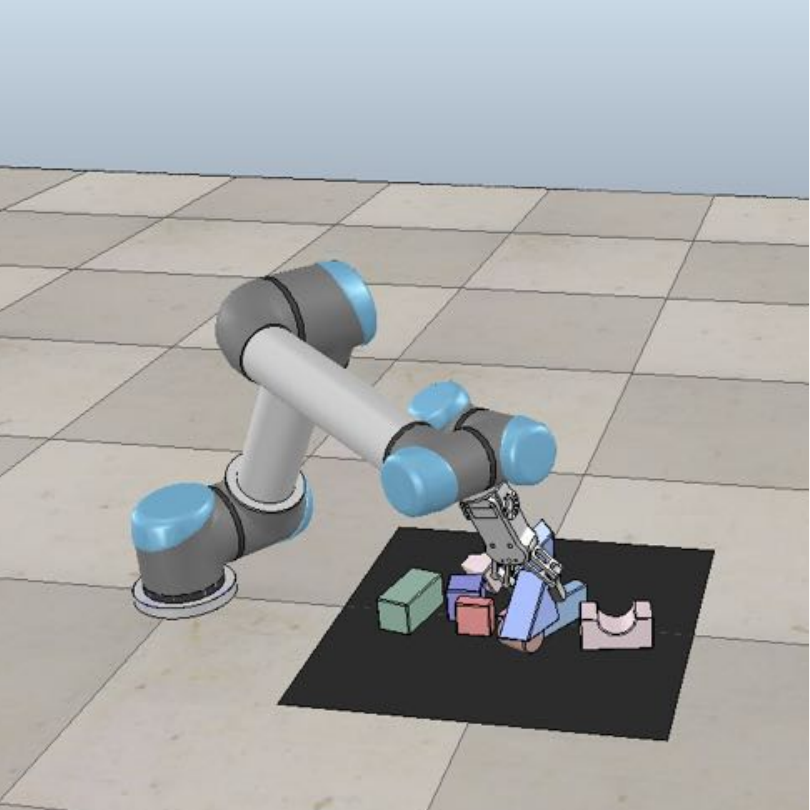}%
        \vspace{-3pt}
       \caption{Standard}
   \end{subfigure}
   \begin{subfigure}{0.15\textwidth}
       \includegraphics[scale = 0.32]{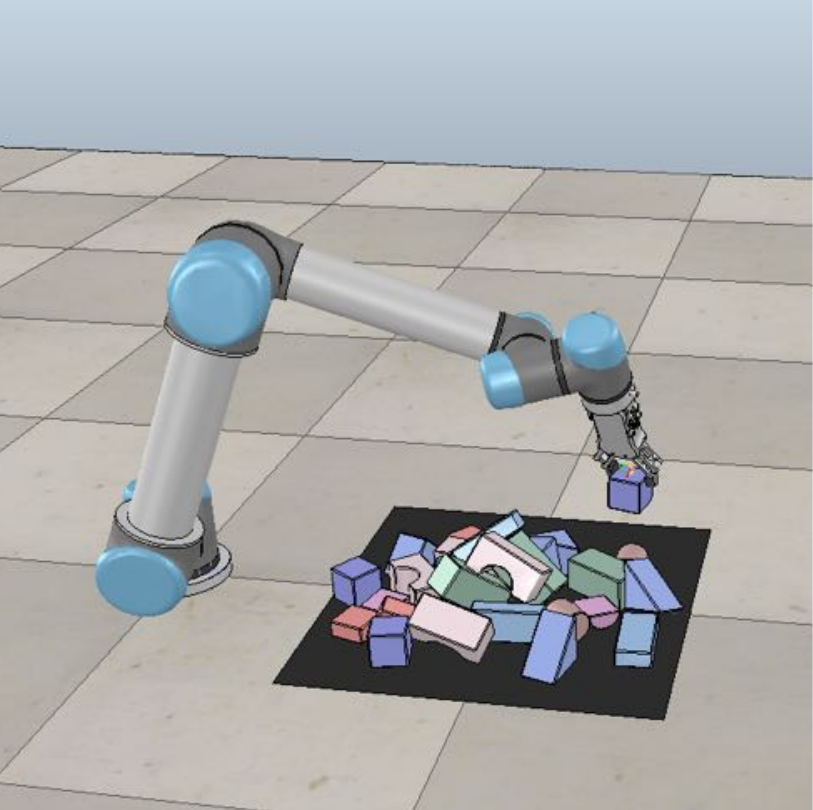}%
        \vspace{-3pt}
       \caption{Challenging}
   \end{subfigure}
   \begin{subfigure}{0.15\textwidth}
       \includegraphics[scale = 0.32]{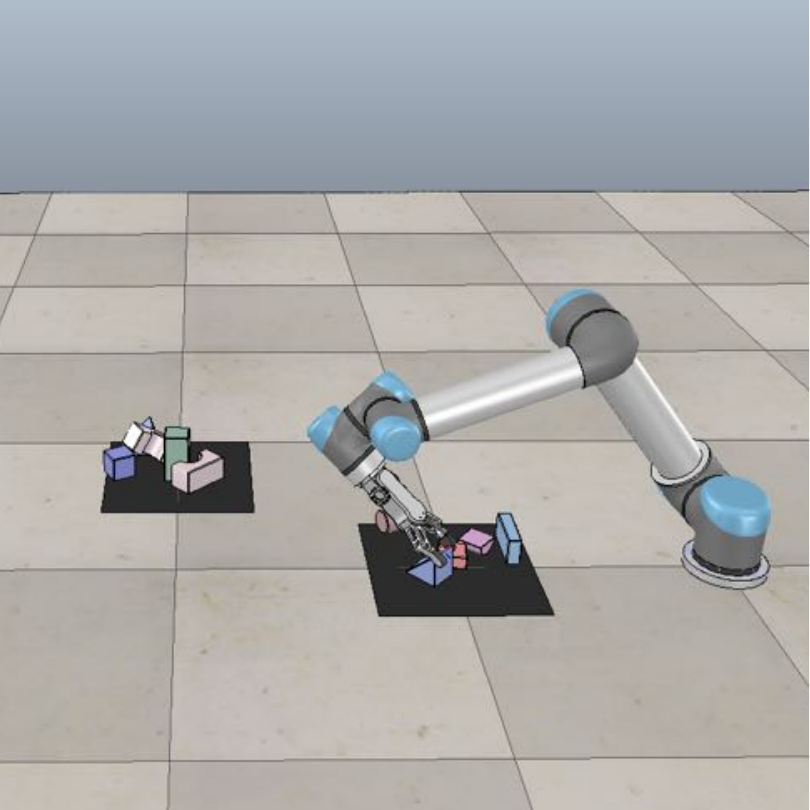}%
        \vspace{-3pt}
       \caption{Ablation}
   \end{subfigure}%
    % \vspace{-5pt}
\caption{\textbf{Simulation experiments.} Our approach is evaluated in (a) the standard scenario with randomly arranged 10 objects, (b) the challenging scenario with 30 randomly dropped objects and (c) ablation study.}
\label{fig:simtests}
\vspace{-8pt}
\end{figure}

\begin{figure}[t]
\centering
   \begin{subfigure}{0.47\textwidth}
       \centering
    \includegraphics[scale=0.58]{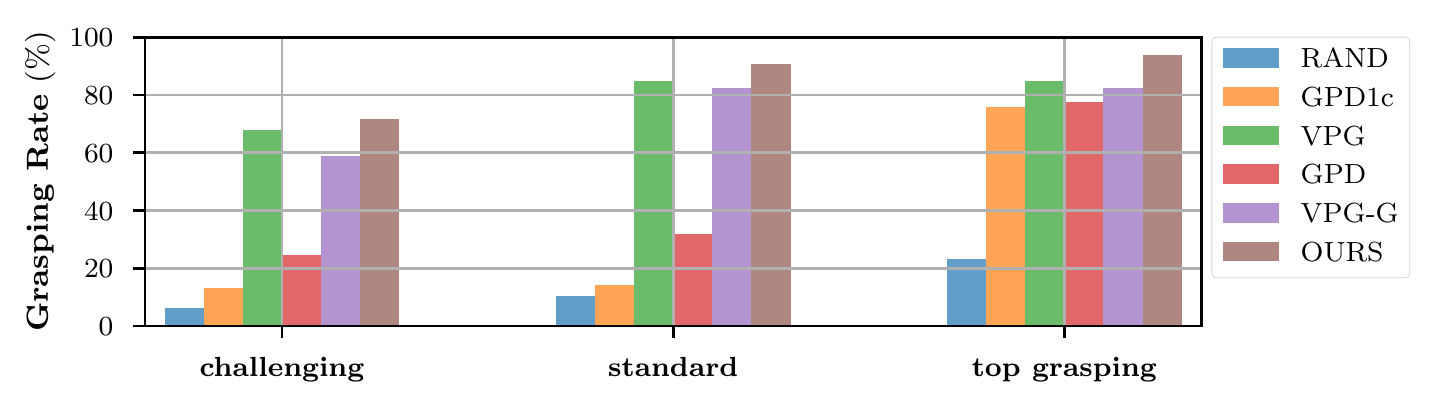}%
    \vspace{-5pt}
   \end{subfigure}
   \begin{subfigure}{0.47\textwidth}
       \centering
    \includegraphics[scale=0.58]{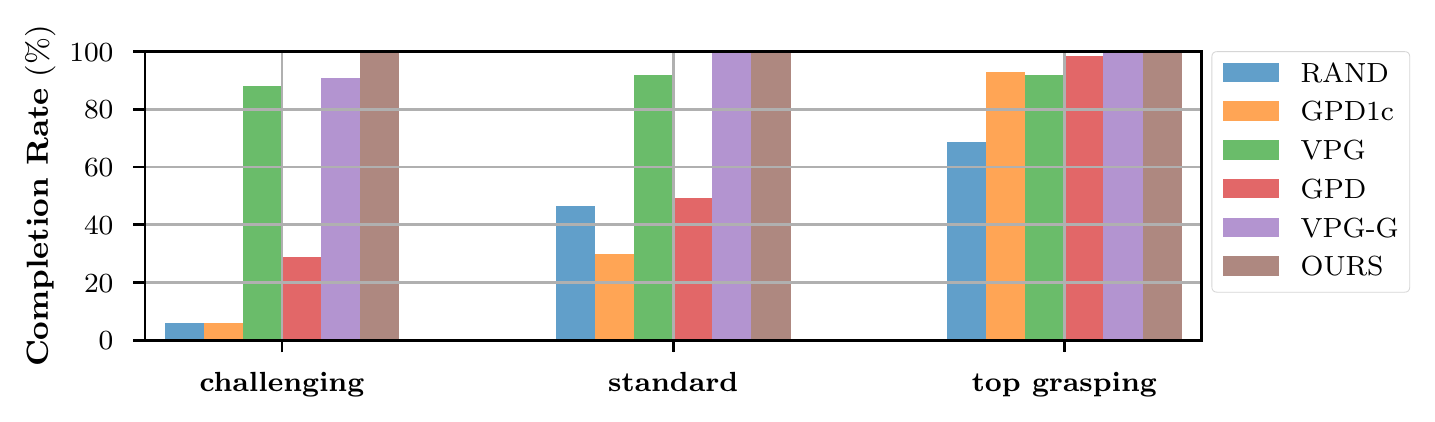}%
    \vspace{-5pt}
   \end{subfigure}
\caption{\textbf{Performance in simulation.} The grasping success rate (top) and completion rate (bottom) of each approach in three different settings. The plots clearly show the effectiveness of our approach, which achieves 71.5\% grasping success rate in challenging scenario and 100\% completion rate in all experiments.}
\label{fig:perf in sim}
\vspace{-8pt}
\end{figure}

\subsection{Simulation Experiments}

We adopt the same testing objects and simulated environment as reported in \cite{zeng2018learning} for a fair comparison. The V-REP \cite{rohmer2013v} scene includes a UR5 robot arm equipped with an RG2 gripper, and the dynamics are simulated with Bullet \cite{Coumans:2015:BPS:2776880.2792704} Physics engine 2.83. We noticed \textbf{\texttt{GPD}} under this one camera setup (\textbf{\texttt{GPD1c}}) can not perform well as their geometric-based grasp candidates search requires a more complete point cloud. We thus added another camera for \textbf{\texttt{GPD}} to reproduce the ``passive point cloud'' setup described in \cite{ten2017grasp}. Gripper grasping point and orientation for each approach are calibrated respectively since their hand configurations are different. The experiments are illustrated in Fig.~\ref{fig:simtests}. For all the experiments, a grasp is successful only if the object is lifted by 15 cm. If the robot failed 10 times (including planning failures) consecutively or all the objects have been removed, this run is completed.

Fig.~\ref{fig:perf in sim} presents the average grasping success rates and completion rates over 30 runs for each method, where grasping success rate $ = \frac{\text{number of successful grasps}}{\text{number of proposed grasps}}$ and completion rate $ = \frac{\text{number of objects grasped}}{\text{total number of objects}}$. We first compare our performance with others in a standard multiple objects grasping scenario. The goal is to grasp 10 objects that are randomly dropped to the center of the ground. We notice that the RP effectively penalizes approaching directions toward the robot when grasping point exceeds its learned thresholds, as shown in Fig.~\ref{fig:rpviz}. To fairly compare our approach and \textbf{\texttt{GPD}} with \textbf{\texttt{VPG}}, we also report the top grasping success rate for each method. An interesting observation is that the performance of \textbf{\texttt{GPD}} improves significantly as top-down grasping poses are mostly stable and reachable in this setting. This indirectly corroborates the importance of reachability awareness when proposing 6-DoF poses. The challenging scenario compares our approach to the others in a densely cluttered setting where 30 objects are randomly dropped to the center. This triples the workspace density thus demonstrates our ability to generalize to more cluttered scenarios. With only one camera and no help of pushing to declutter the challenging scenario, our approach achieved the highest grasping success rate and completion rate. Our approach suggests feasible grasp poses as long as the object is within the view, while \textbf{\texttt{VPG}} may accidentally force objects out of its workspace.

\begin{table}[t]
\caption{Ablation Study of Reachability Predictor}
\label{tab:PR}
\begin{center}
\begin{tabular}{|c||c||c||c|}
\hline
  & RAND & 3DCNN & OURS\\
\hline
Grasping Success Rate & 13.33 & 32.67 & \textbf{82.67}\\
\hline
Planning Success Rate & 26.67 & 37.33 & \textbf{96.00}\\
\hline
\end{tabular}
\vspace{-8pt}
\end{center}
\end{table}

We report an ablation study of the reachability predictor by grasping from two clusters of objects, one of which is partially unreachable. We tested our approach with (\textbf{\texttt{OURS}}) and without (\textbf{\texttt{3DCNN}}) the RP to grasp the objects. Given only 5 chances for each run, the robot has to choose the most stable and reachable pose to succeed. We report the average grasping success rate and planning rate over 30 runs for \textbf{\texttt{RAND}}, \textbf{\texttt{3DCNN}} and \textbf{\texttt{OURS}} in Table I, for a total of 150 grasps. Planning efficiency $ = \frac{\text{number of successful planning}}{\text{total number of grasping planned}}$. Our reachability aware 3D CNN is able to achieve 96\% planning rate, an improvement of 58.67\% compared to 3D CNN only.

\begin{table}[b]
\caption{Challenging Scenario in Real-world}
\label{tab:challenging}
\begin{center}
\begin{tabular}{|c||c||c||c||c|}
\hline
  & RAND & GPD1c & VPG & OURS\\
\hline
Grasping Success Rate & 13.10 & 31.23 & 64.76 & \textbf{75.20}\\
\hline
Completion Rate & 14.58 & 12.75 & 94.33 & \textbf{100.0}\\
\hline
\end{tabular}
\vspace{-8pt}
\end{center}
\end{table}

\begin{figure}[t]
\centering
    \includegraphics[scale=0.34]{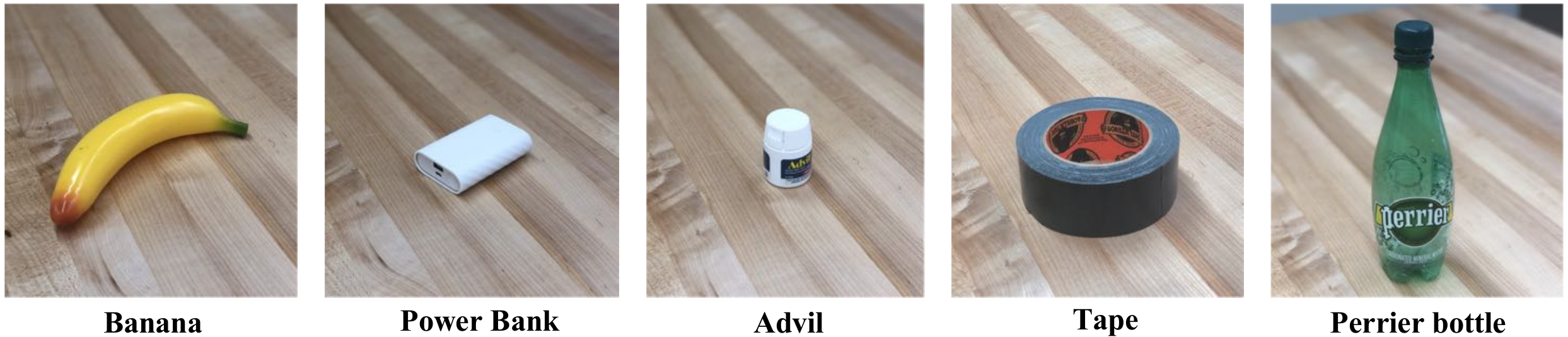}%
    \vspace{-4pt}
\caption{\textbf{Real-world novel objects.} We use five novel objects to show the generalization capability of our approach. The shape, texture and size of the testing objects are different from the training objects in simulation as well.}
\label{fig:novelobj}
\vspace{-8pt}
\end{figure}

\begin{figure}[t]
\centering
    \includegraphics[scale=0.57]{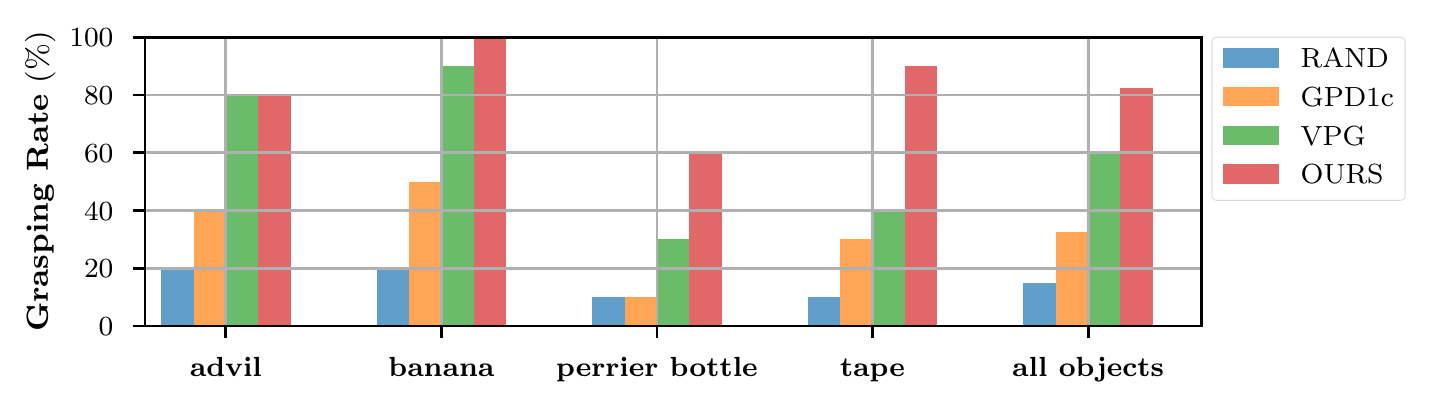}%
    \vspace{-5pt}
\caption{
%\todo{Inconsistent names: `GPD-1c' and `GPD1c'. Please stick to one name.}
\textbf{Grasping generalizability.} The grasping success rate of four approaches on the 5 test objects. The average grasping success rate of our approach is 82.5\%, surpassing all others on novel objects.}
\label{fig:novel perf}
\vspace{-5pt}
\end{figure}

\begin{figure}[t]
\centering
    \includegraphics[scale=0.31]{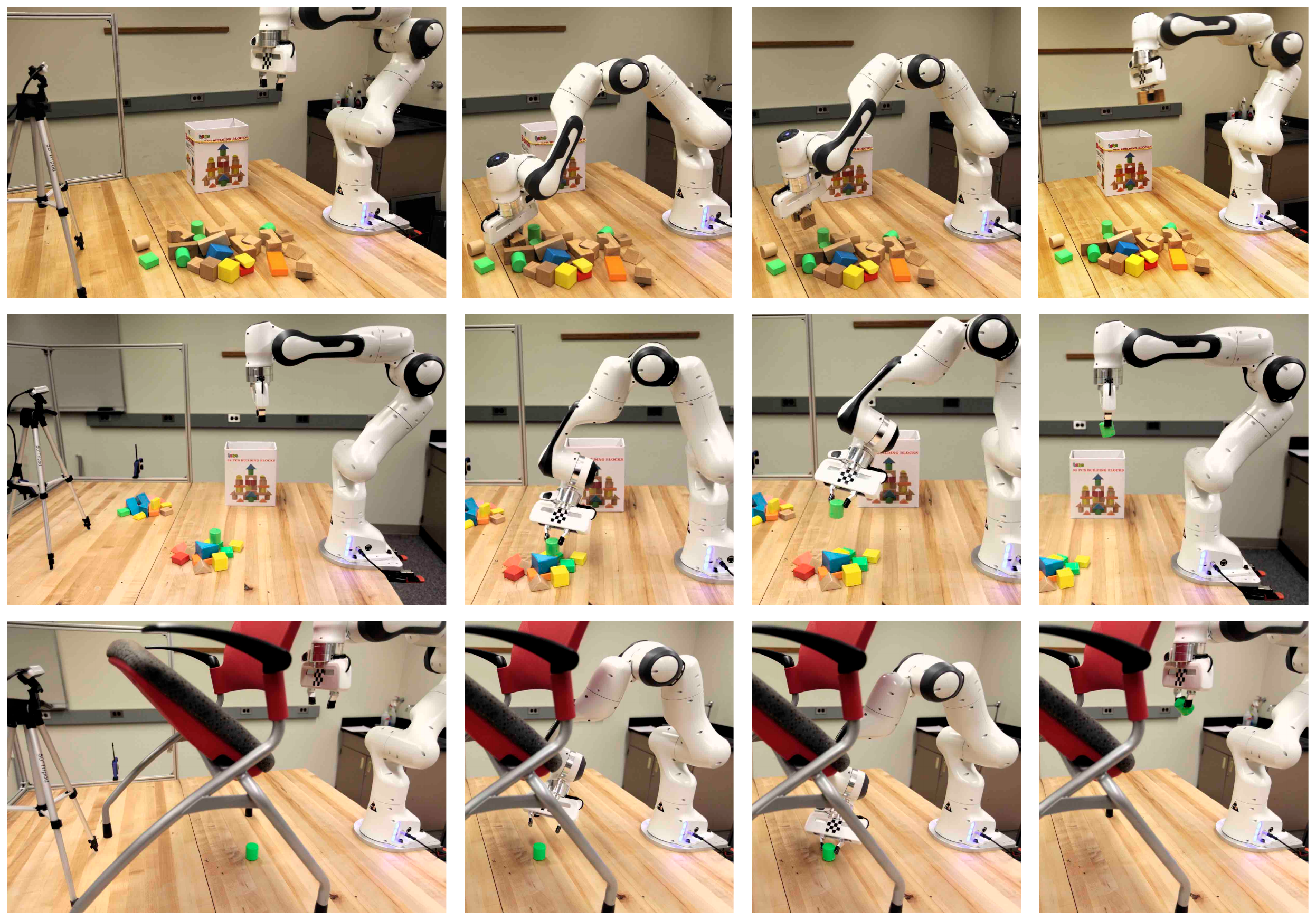}%
    \vspace{-5pt}
\caption{\textbf{Real-world experiment pictures.} The voxel-based 3D CNN approach minimizes the gap between simulation and real-world. Our simulation trained network is able to clean dense clutters (top), differentiate reachable grasps (middle) and grasp an object in a constrained environment which is not feasible by the top grasping approaches, such as VPG (bottom).}
\label{fig:realtests}
\end{figure}

\subsection{Real Robot Experiments}

We evaluate our approach on a Franka Emika Panda robot arm without any fine-tuning. Fig.~\ref{fig:realtests} shows the real robot experiments. We use 30 toy blocks to reproduce the simulated challenging scenario and ablation study, five random household objects to test the generalization capability and a chair positioned over the object to show the flexibility of our approach. 

We compare our approach with \textbf{\texttt{RAND}}, \textbf{\texttt{GPD}}, and \textbf{\texttt{VPG}} in the challenging scenario. Due to hardware limitations, we were not able to implement \textbf{\texttt{GPD}} with two cameras. Table II compiles the performance of each method from averaging the results of 10 runs. Our approach is able to perform consistently in real-world and achieves the best grasping success rate and completion rate. Despite the robustness of 3D CNN to sensor noise \cite{choi2018learning}, our grasp sampling algorithm may propose a vacant grasping point that contributes to a misaligned pose. \textbf{\texttt{VPG}} performs efficiently as long as the object is not forced out of the constrained workspace. \textbf{\texttt{GPD1c}} suffers from predicting unreachable poses and incomplete point cloud.

In real robot ablation study, our approach demonstrates an efficiency improvement as seen previously, but the gap between the simulated UR5 and the real Panda arm deteriorates the performance. Quantitative results are shown in Table III.

\begin{table}[t]
\caption{Reachability Predictor in Real-world}
\label{tab:pr_real}
\begin{center}
\begin{tabular}{|c||c||c||c|}
\hline
  & RAND & 3DCNN & OURS\\
\hline
Grasping Success Rate & 3.33 & 23.33 & \textbf{66.67}\\
\hline
Planning Efficiency & 26.67 & 34.67 & \textbf{88.67}\\
\hline
\end{tabular}
\vspace{-8pt}
\end{center}
\end{table}

To test our real-world generalizability, we further experiment with objects that have never been seen during training, shown in Fig.~\ref{fig:novelobj}. We ran 10 tests on each object, the result is summarized in Fig.~\ref{fig:novel perf}. Our approach is able to extract 3D geometric features from novel shapes and select grasp pose candidates accordingly. We noticed that when grasping considerably larger objects such as the Perrier bottle, our approach prefers smaller shapes such as the bottle neck and cap. This is due to the limited physical voxel grid size that can only fit in shapes that are similar in size to our training objects. We also give an example of our system's grasping flexibility by positioning an object under a chair to reflect real-world challenges. As shown in Fig.~\ref{fig:realtests}, the robot is able to complete the task by selecting a feasible grasp which is not possible with the top grasping approaches such as \textbf{\texttt{VPG}}.

\section{CONCLUSIONS}

In this work, we presented a deep learning approach to generate 6-DoF grasp poses with reachability awareness. A 3D CNN  model that estimates grasping stability was trained with a large-scale dataset obtained from simulated self-supervision. A reachability predictor that improves reachability awareness of 3D CNN was trained similarly. We outperformed several comparable deep learning approaches in both simulation and real-world. Furthermore, our method achieved 82.5\% grasping success rate on unknown objects. Ablation study showed the RP significantly increased the planning efficiency of 3D CNN by 54\% in real-world experiments.

The limitations of our work suggest two directions for future work. First, our sampling algorithm selects grasping points from the point cloud, which is susceptible to sensor noise. It is of our interest to investigate whether a voxel-based method is able to enhance its robustness. Second, we only applied reachability predictor to grasping and would like to explore if RP can improve other manipulation tasks.


%%%%%%%%%%%%%%%%%%%%%%%%%%%%%%%%%%%%%%%%%%%%%%%%%%%%%%%%%%%%%%%%%%%%%%%%%%%%%%%%
\bibliographystyle{IEEEtran}
\bibliography{IEEEabrv,IEEEexample}
\end{document}